\documentclass[conference]{IEEEtran}
\IEEEoverridecommandlockouts

\usepackage{amssymb}
\usepackage{cite}
\usepackage{amsmath,amssymb,amsfonts}
\usepackage{algorithmic}
\usepackage{graphicx}
\usepackage{textcomp}
\usepackage{xcolor}
\usepackage{booktabs}
\def\BibTeX{{\rm B\kern-.05em{\sc i\kern-.025em b}\kern-.08em
    T\kern-.1667em\lower.7ex\hbox{E}\kern-.125emX}}

\makeatletter

\newcommand{\Rmnum}[1]{\expandafter\@slowromancap\romannumeral #1@}
\makeatother

\makeatletter
\newcommand{\linebreakand}{%
  \end{@IEEEauthorhalign}
  \hfill\mbox{}\par
  \mbox{}\hfill\begin{@IEEEauthorhalign}
}
\makeatother
    
\begin{document}

\title{\title{Non-stationary BERT: Exploring Augmented IMU Data For Robust Human Activity Recognition
}
}

\author{\IEEEauthorblockN{1\textsuperscript{st} Ning Sun}
\IEEEauthorblockA{ 
\textit{The University of Hong Kong}\\
\textit{OPPO Research Institute}\\
Hong Kong, China \\
sn0923@connect.hku.hk 
}

\and

\IEEEauthorblockN{1\textsuperscript{st} Yufei Wang}
\IEEEauthorblockA{
\textit{Shanghai Jiao Tong University}\\
Shanghai, China \\
arthur-w@sjtu.edu.cn}

\and

\IEEEauthorblockN{3\textsuperscript{rd} Yuwei Zhang}
\IEEEauthorblockA{
\textit{OPPO Research Institute}\\
Shanghai, China \\
zhangyuwei2@oppo.com
}

\linebreakand

\\
\IEEEauthorblockN{4\textsuperscript{rd} Jixiang Wan}
\IEEEauthorblockA{
\textit{OPPO Research Institute}\\
Shanghai, China \\
wanjixiang@oppo.com}

\and

\\
\IEEEauthorblockN{4\textsuperscript{rd} Shenyue Wang}
\IEEEauthorblockA{
\textit{OPPO Research Institute}\\
Shanghai, China \\
wangshenyue@oppo.com}

\and

\\
\IEEEauthorblockN{4\textsuperscript{rd} Ping Liu}
\IEEEauthorblockA{
\textit{OPPO Research Institute}\\
Shanghai, China \\
liuping1@oppo.com}

\and

\\
\IEEEauthorblockN{4\textsuperscript{rd} Xudong Zhang}
\IEEEauthorblockA{
\textit{OPPO Research Institute}\\
Shanghai, China \\
zhangxudong@oppo.com}
}

\maketitle

\begin{abstract}
Human Activity Recognition (HAR) has gained great attention from researchers due to the popularity of mobile devices and the need to observe users' daily activity data for better human-computer interaction.
In this work, we collect a human activity recognition dataset called OPPOHAR consisting of phone IMU data. 
To facilitate the employment of HAR system in mobile phone and to achieve user-specific activity recognition, we propose a novel light-weight network called Non-stationary BERT with a two-stage training method. We also propose a simple yet effective data augmentation method to explore the deeper relationship between the accelerator and gyroscope data from the IMU. The network achieves the state-of-the-art performance testing on various activity recognition datasets and the data augmentation method demonstrates its wide applicability.
\end{abstract}


\section{Introduction}
Human activity recognition (HAR) has been an important research area for decades and plays a crucial role in many applications, such as human-computer interaction, human behaviour analysis, and ubiquitous computing ~\cite{wan2020deep,lentzas2020non,aziz2023real}. 
In recent years, advancements in sensing analytics of mobile devices have driven the rapid development of human activity recognition. They provide opportunities for continuous tracking of physiological signals and boost seamless communication between humans and machines~\cite{garcia2020public}.
Inertial measurement units (IMUs), which contain accelerometers and gyroscopes, are typically electromechanical or solid-state devices that detect linear acceleration and angular velocity~\cite{samatas2022inertial}. 
They are widely applied in mobile devices due to their adaptability and simplicity~\cite{ni2024survey}.

By utilizing the data from the IMU implemented in mobile devices, real-time human activity recognition becomes feasible. 
Due to the power and computing limitations of mobile devices, classifiers for HAR are typically lightweight. Various Machine Learning (ML) algorithms, RNN, CNN, Transformer, and mixed-architecture models are proposed~\cite{nia2023human,hou2020study,agarwal2020lightweight,ankita2021efficient,coelho2021lightweight,zhou2022tinyhar}. 

In alignment with the development of HAR systems, several datasets are introduced, including UCI~\cite{reyes2016transition}, shoiab ~\cite{shoaib2014fusion}, mhealth ~\cite{oresti2014mhealthdroid}.
Prior studies~\cite{choi2023effects,xu2023practically,qian2022makes} also apply data augmentation methods on IMU data, e.g., adding random noise, rotation, flipping, bias, physical transformation, etc. These data augmentation methods increase the diversity of the training data, thereby mitigating over-fitting and improving the generalization ability of the HAR model.

However, the prior works have several drawbacks.
Firstly, existing models frequently struggle to adapt to real-world scenarios as each user has his own movement pattern. A user might want to record some specific but uncommon activity, such as skiing. Previous models need to be trained from scratch to accommodate a new activity. 
Secondly, they do not delve deeply enough into the relationship between accelerator and gyroscope data for a robust data augmentation method. 

To address these drawbacks, we implement a two-stage inference pipeline. As depicted in the Figure \ref{fig:NSO}, the model separates the pretraining of the Encoder and the finetuning of the Classifier. Pretraining always requires a significant amount of time, thus can be done in the cloud; Finetuning can be conducted on users' mobile devices as we want to use the user-labeled activity data to train their own classifiers, which also ensures user privacy. 

In this work, we have three primary contributions:
\begin{itemize}
    \item We propose a new human activity recognition dataset named OPPOHAR, which encompasses a diverse range of human activities using the phone in various gestures.
    \item We introduce an effective data augmentation method tailored for processing IMU data.
    \item We propose a lightweight network optimized for distributed deployment for HAR, dedicated to privacy-protecting user-specific activity recognition.
\end{itemize}

\section{Dataset}
\subsection{Common Activities}
We define seven common human activities for mobile devices: staying still, walking, walking up and down stairs, running, cycling, taking car, and taking subway. For convenience, we name them respectively as activity A, B, C, D, E, F, and G. These activities cover a large variety of scenarios with phones in human daily life, making the dataset practically valuable. Data are collected from seven different activities mentioned above by two collectors. Take the activity walking as an example, we use 2 different devices and different hands holding the phone to collect data at 3 different speeds, slow, medium, and fast. We illustrate the duration of time for each activity in our dataset as in the first two rows of Table~\ref{tab:oppodataset}. 

\subsection{Uncommon Activities}

We also construct an uncommon dataset including activities distinguished from daily activities. We define six distinct hand gestures holding mobile devices: rotating the phone like a circle, tracing a "W" shape in the air, tracing a "Z" shape in the air, vigorously shaking the phone up and down, tapping the phone from behind, gently shaking the phone from side to side. For convenience, we name them respectively as activity a, b, c, d, e, and f. These activities deviate from typical phone usage behaviors and serve as triggers for specialized phone functions, enabling quick access to certain functions of their devices. 
Each activity mentioned above is recorded in ten sets collected by three collectors.
Within each set, the activity will be repeated continuously approximately 20 times. In the last two rows of Table~\ref{tab:oppodataset}, we illustrate the duration of time for each activity in our dataset. 

\begin{table}[htbp]
\centering
\caption{Illustration of duration of each activities in OPPOHAR.}
\label{tab:oppodataset}
\resizebox{1\linewidth}{!}{
\begin{tabular}{|c|c|c|c|c|c|c|c|}
\hline
Activity &  A &  B & C & D & E & F & G \\ 
\hline
Duration(s)   & 18570  & 10806 & 8442 & 1998 & 16428 & 14694 & 5010  \\ 
\hline
Activity & a  &  b & c & d & e & f & \\ 
\hline
Duration(s)   & 804  & 1602 & 1152 & 1614 & 942 & 1194 &  \\ \hline
\end{tabular}
}
\end{table}



\section{Methodology}
\begin{figure}
    \centering
    \includegraphics[width=0.9\linewidth]{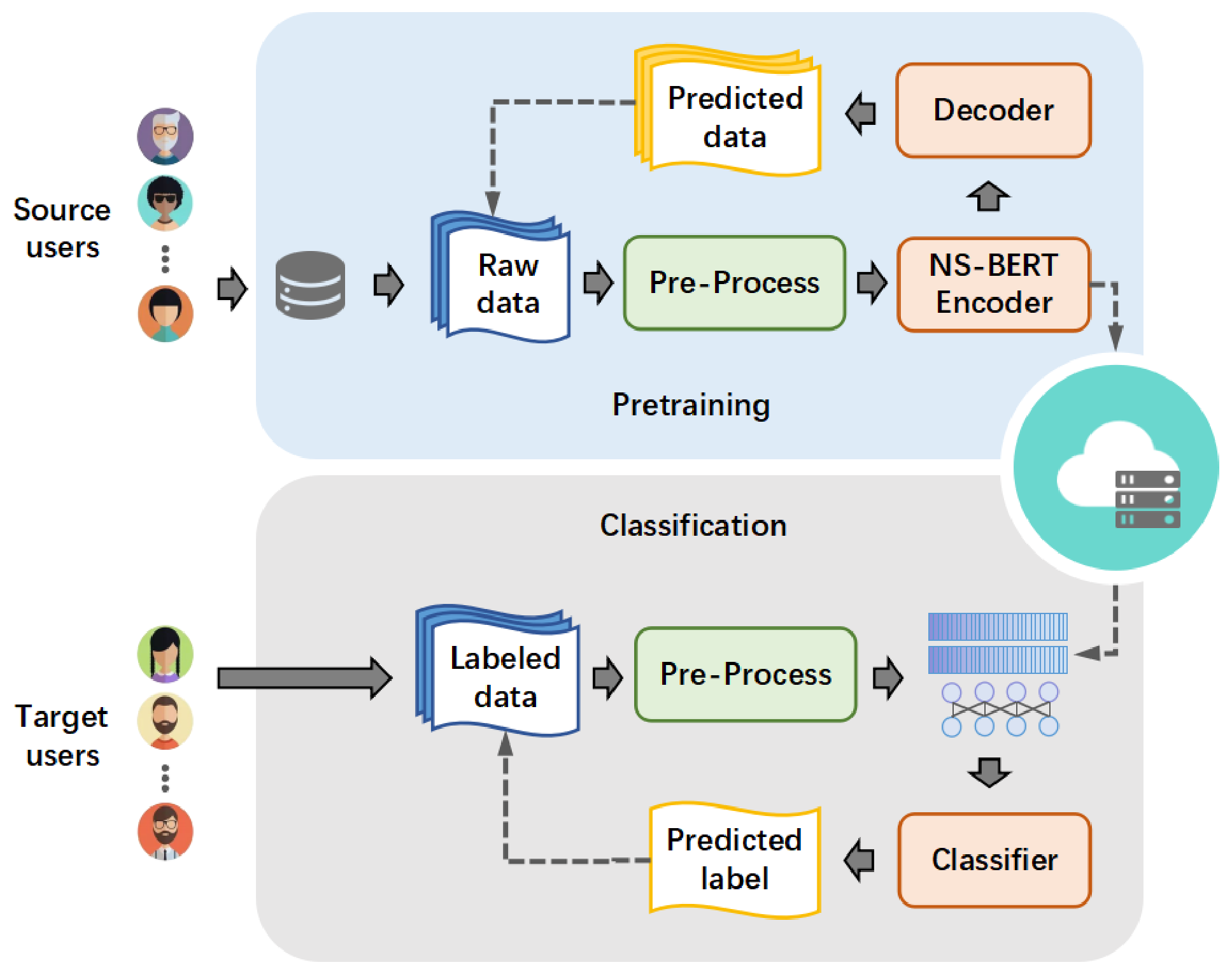}
    \caption{The Overview of our Non-stationary BERT network}
    \label{fig:NSO}
\end{figure}

In this section, we present the overall design for our Non-stationary BERT (Bidirectional Encoder Representations from Transformers~\cite{kenton2019bert}) network. 
As depicted in Figure \ref{fig:NSO}, it includes two phases: self-supervised pretraining and supervised classifying. Inspired by the previous work LIMU-BERT~\cite{xu2021limu}, we adopt a similar architecture, which contains three parts: a BERT-like Encoder, a Decoder, and a Classifier. 

During the self-supervised pretraining phase, raw IMU data is first pre-processed and then sent to the Encoder-Decoder module for sequence recovery task. The NS-BERT Encoder is designed to learn an effective hidden representation of the input sequence and the Decoder is designed to recover this time sequence based on former hidden representation. 
For the supervised classification phase, we freeze the Encoder and replace the Decoder with a Classifier for human activity recognition. The network is finetuned for the HAR task. 
The pretraining network can be trained online and later send Encoder to all users. Classifiers can be trained on user mobile devices. 

\subsection{Data Augmentation}
\label{sec:FM}
The accelerator data and the gyroscope data from IMU present different characteristics of the activity. Previous work ~\cite{li2021data,tran2020data,jaafer2020data,xu2023practically} do not consider the inter-relationship between them. They usually concatenate them and send them directly to the network, or use two encoders to extract the features respectively and concatenate them with later fusion. 

Instead, we propose an efficient data augmentation method, FM (Factorization Machine~\cite{rendle2010factorization}).
Consider a time sequence from IMU data, $acc_{i,j}$ and $gyro_{i,k}$ represent respectively the accelerator data and gyroscope data of time step i and of axis j, where j and k come from axis x or y or z. We select a sequence of $acc_{i,j}$ and a sequence of $gyro_{i,k}$ and multiply them at each time step and get a new sequence called $fm_{i,j,k}$ as in equation below: 
\begin{equation}
fm_{i,j,k} =  acc_{i,j} \times gyro_{i, k}
\end{equation}

As the accelerator and gyroscope have each 3 axes, we have a total of 9 new time sequences, which are also regarded as the input. These created feature sequences contain hidden trends of data fluctuation, facilitating the neural network to discover the underlying laws between them. 




With a detection window of 6 seconds, the raw IMU data is split into sequences. After data normalization, we got data in a shape like $T * SR * FS$, where T, SR, and FS denote time length, sampling rate, and feature shape respectively. In this scenario, the feature shape is 15. In the following Embedding phase, we project feature shape from 15 into embedding size.


\subsection{Model Architecture}
\subsubsection{Encoder}

\begin{figure}[th]
    \centering
    \includegraphics[width=0.75\linewidth]{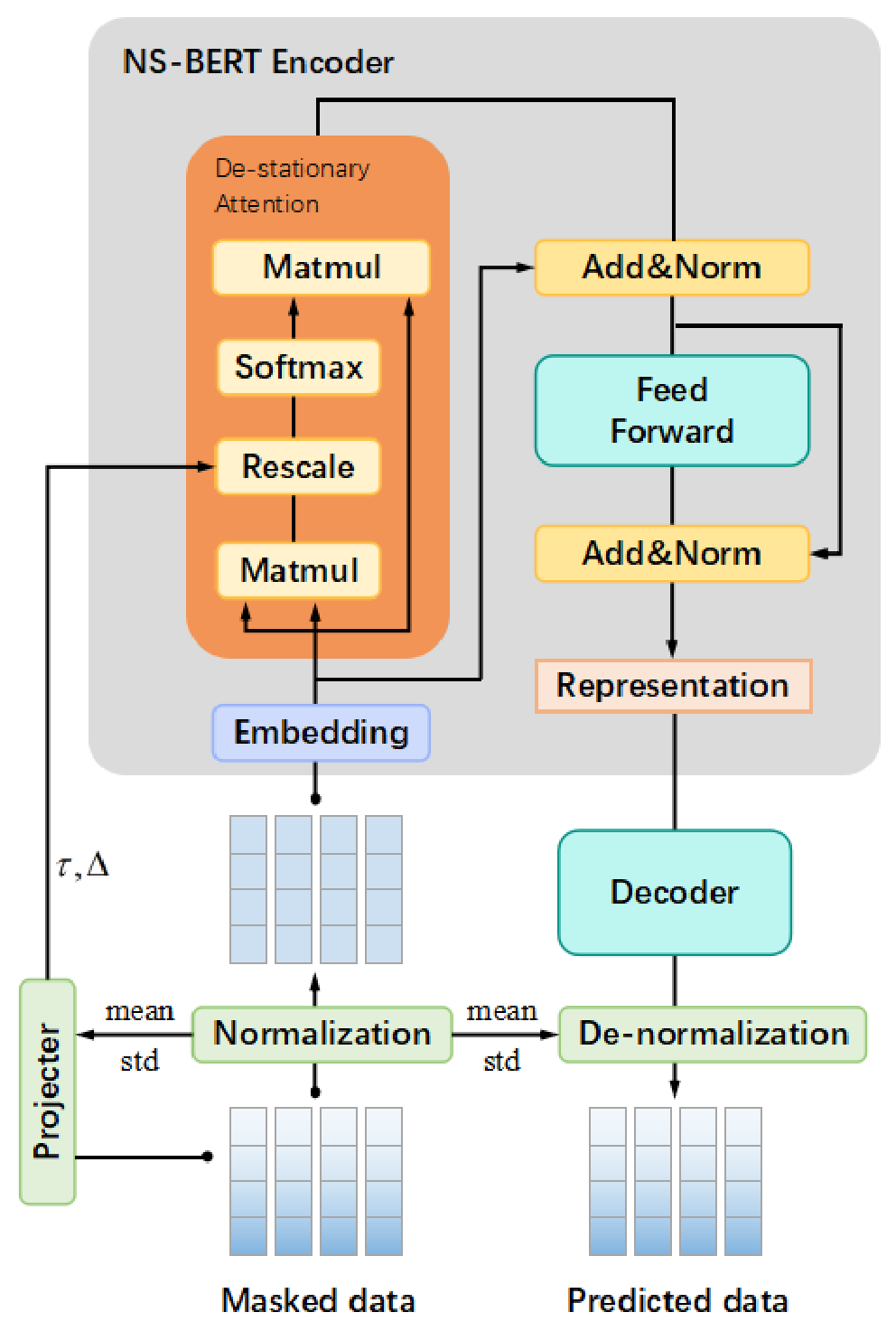}
    \caption{Non-stationary BERT pretraining workflow}
    \label{fig:NS_6}
\end{figure}

Time series is commonly non-stationary, which means that the statistical properties and joint distribution of time series can change over time. 
For IMU data of HAR, its statistical properties can change drastically when the users change the posture of the device or their state of motion. Eliminating the impact of such events can help improve predictability. 
Previous works~\cite{shen2023gbt,wan2024tcdformer} implement stationarization to handle time series prediction tasks. 
Borrowed from the idea of Non-stationary Transformer~\cite{liu2022non}, instead of using self-attention~\cite{vaswani2017attention}, we pioneeringly apply the Series Stationarization and De-stationary Attention operations to a BERT-like network to handle IMU data, which enhances predictability of the series and keeps non-stationary information of the original series simultaneously.

The Series Stationarization module includes the Normalization and De-normalization modules as shown in Figure~\ref{fig:NS_6}. 
For each input batch, the Normalization module of NS-BERT encoder records the mean $\mu_{X}$, and the standard deviation $\sigma_{X}$ of each sequence $X =[x_{1},x_{2},\cdots,x_{S}]^{T}\in\mathbb{R}^{S \times E}$ . 
S denotes sequence length and E denotes the number of features. 


After the normalization, the input is more stationary and predictable. Thus the encoder input turns from $Q,K,V$ to $Q',K',V'$, and they have such relationship:
\begin{equation}
\begin{aligned}
    Q'K'^{T}& =\frac{(Q-\mathbf{1}\mu_{Q}^{T})(K-\mathbf{1}\mu_{K}^{T})^{T}}{\sigma_{Q}\sigma_{K}} \\
    & = \frac{1}{\sigma_{X}^{2}}(QK^{T}-\mathbf{1}(\mu_{Q}^{T}K^{T})-(Q\mu_{K})\mathbf{1}^{T}+\mathbf{1}{\mu_{Q}^{T}\mu_{K}}\mathbf{1}^{T})
\end{aligned}
\end{equation}

To recover the non-stationary information, De-stationary Attention approximates the original attention by introducing two de-stationary factors. In the following calculation, we regard $\sigma_{X}^{2}$ as $\tau$ and $\mu_{Q}^{T}K^{T}$ as $\Delta$. The original attention score without normalization can be deduced as follows:
\begin{equation} 
\begin{aligned} 
    Attention(Q,K,V) & = Softmax(\frac{QK^{T}}{\sqrt{d_{k}}})V  \\
    & = Softmax(\frac{\tau Q'K'^{T}+\Delta}{\sqrt{d_{k}}})V
\end{aligned}
\end{equation}

Two de-stationary factors are learned by MLP using $\mu_{X}$, $\sigma_{X}$ and raw data. By introducing these factors, we can recover the non-stationary information deep in the model, which improves its performance in predicting real-world time series. 

\subsubsection{Decoder}
The task of the Decoder is to reconstruct the original values of IMU sequences with the representations extracted by Encoder. During the training period, we use a Mean Square Error loss to optimize. 

At the output side of decoder, $\mu_{X}$ and $\sigma_{X}$ are used to transform the model output, $y'=[y_{1}',y_{2}',\cdots,y_{S}']^{T}\in\mathbb{R}^{S \times E}$. De-normalization module operates as follows:
\begin{equation}
    \hat{y}_{i} = \sigma_{X}\odot y_{i}'+\mu_{X}
\end{equation}



\section{Experiment and Results}
\begin{table*}[ht]
\centering
\caption{Performance comparison on original data}
\label{tab:performance_ori}
\resizebox{0.7\linewidth}{!}{
\begin{tabular}{lcccccccccc}
\toprule 
Dataset & \multicolumn{2}{c}{OPPOHAR} & \multicolumn{2}{c}{UCI} & \multicolumn{2}{c}{Mhealth}  & \multicolumn{2}{c}{Shoaib}  \\
\midrule
Metric & Acc & F1 & Acc & F1 & Acc & F1 & Acc & F1  \\
\midrule 
NS-GRU & \textbf{0.921} & \textbf{0.920} & \textbf{0.962} & \textbf{0.966} & \textbf{0.890} & \textbf{0.889} & \textbf{0.975} & \textbf{0.974} \\
LIMU-GRU~\cite{xu2021limu} & 0.891 & 0.893 & 0.924 & 0.927 & 0.882 & 0.800 & 0.966 & 0.965 \\
GRU~\cite{chung2014empirical} & 0.767 & 0.788 & 0.929 & 0.932 & 0.706  & 0.670 & 0.969 & 0.968 \\
DCNN~\cite{yang2015deep} & 0.871 & 0.850 & 0.925 & 0.922 & 0.845 & 0.797 & 0.930 & 0.928 \\
Deepsense~\cite{yao2017deepsense} & 0.735 & 0.724 & 0.748  & 0.725  & 0.537 & 0.526 & 0.805 & 0.798 \\
LSTM~\cite{shi2015convolutional} & 0.785 & 0.810 & 0.910 & 0.907 & 0.697 & 0.637 & 0.950 & 0.949 \\
\bottomrule 
\end{tabular}
}
\end{table*}

\begin{table*}[ht]
\centering
\caption{Performance comparison on augmented data}
\label{tab:performance_aug}
\resizebox{0.7\linewidth}{!}{
\begin{tabular}{lcccccccccc}
\toprule 
Dataset & \multicolumn{2}{c}{OPPOHAR} & \multicolumn{2}{c}{UCI} & \multicolumn{2}{c}{Mhealth}  & \multicolumn{2}{c}{Shoaib}  \\
\midrule
Metric & Acc & F1 & Acc & F1 & Acc & F1 & Acc & F1  \\
\midrule 
NS-GRU-FM & \textbf{0.969} & \textbf{0.970} & \textbf{0.986} & \textbf{0.986} & \textbf{0.956} & \textbf{0.956} & \textbf{0.987} & \textbf{0.986} \\
NS-GRU & 0.921 & 0.920 & 0.962 & 0.966 & 0.890 & 0.889 & 0.975 & 0.974 \\
\midrule 
LIMU-GRU-FM & \textbf{0.893} & 0.886 & \textbf{0.952} & \textbf{0.954} & \textbf{0.949} & \textbf{0.945} & \textbf{0.985} & \textbf{0.985} \\
LIMU-GRU~\cite{xu2021limu} & 0.891 & \textbf{0.893} & 0.924 & 0.927 & 0.882 & 0.800 & 0.966 & 0.965 \\
\midrule 
GRU-FM & \textbf{0.805} & \textbf{0.829} & \textbf{0.957}  & \textbf{0.960}  & \textbf{0.809} & \textbf{0.760} & \textbf{0.957} & \textbf{0.960} \\
GRU~\cite{chung2014empirical} & 0.767 & 0.788 & 0.929 & 0.932 & 0.706 & 0.670 & 0.929 & 0.932 \\
\bottomrule 
\end{tabular}
}
\end{table*}

\subsection{Evaluation}
\subsubsection{Datasets and Models}
We select our OPPOHAR and another three public datasets, including UCI, Mhealth, and Shoaib for evaluation. We choose accuracy and F1-score as our metrics.
All datasets are equally down-sampled to 20Hz and sliced into non-overlapping windows with a length of 120 units, which is a time span of 6 seconds. 

For comparison with our NS-BERT, we choose LIMU-BERT~\cite{xu2021limu}, DCNN~\cite{yang2015deep}, Deepsense~\cite{yao2017deepsense}, 
GRU~\cite{chung2014empirical} and LSTM~\cite{shi2015convolutional}. 
For NS-BERT and LIMU-BERT, they both adopt a two-stage training method. The preliminary experiments show that the GRU classifier achieves the best results for finetuning them. Thus, in subsequent experiments, we rename NS-BERT as NS-GRU, indicating we use GRU for the classifier, the same as LIMU-GRU.


\subsubsection{Experiment Settings}
To ensure the experiment's fairness and prevent over-fitting, different training epochs are chosen according to the size of the datasets. 
For LIMU-BERT and NS-BERT, the number of pretraining epochs are 1500-5000. 
The classification training adopts 1000-1200 epochs. All models utilize the same dataset for training, validation, and testing. 
The learning rate in the pretraining phase and classification phase are 0.0001 and 0.001, respectively. The batch size is 2048.

\subsubsection{Model Performance} 
In Table~\ref{tab:performance_ori}, our baseline model achieves the best performance among all models. 
Compared with the baseline LIMU-GRU, which does not have stationarization mechanism, our model NS-GRU shows a great performance improvement, proving the effectiveness of Non-stationary structure. 


\subsection{Ablation Study}
\subsubsection{Model Performance on Augmented Data}
To evaluate the effectiveness of the data augmentation method we presented in~\ref{sec:FM}, we compare the accuracy of models with and without data augmentation. For simplicity, we only compare the models whose classifier phase uses GRU. Table~\ref{tab:performance_aug} presents the performance of NS-GRU, LIMU-GRU, and GRU on four datasets before and after data augmentation. ``FM'' denotes using data augmentation method. The number in bold indicates higher performances.
All models' performance has greatly improved with the data augmentation method. According to our research, our model NS-GRU-FM achieves the state-of-the-art performance on UCI and OPPOHAR datasets in current open-source methods.



\subsubsection{Model Performance on User-Specific Data}
To facilitate user-specific activity recognition, we pretrain NS-GRU on common activities and finetune on uncommon activities as all other models. As shown in Table~\ref{tab:performance_usrspe}, our model achieves the best results among all models, suggesting the effectiveness of the pretrained Encoder for special action recognition. 


\begin{table}[ht]
\centering
\caption{Performance comparison on user-specific data}
\label{tab:performance_usrspe}
\resizebox{0.85\linewidth}{!}{
\begin{tabular}{lcccc}
\toprule 
Data Type & \multicolumn{2}{c}{Original} & \multicolumn{2}{c}{FM-Augmented} \\
\midrule
Metric & Acc & F1 & Acc & F1 \\
\midrule 
NS-GRU & \textbf{0.930} & \textbf{0.931} & \textbf{0.982} & \textbf{0.984} \\
LIMU-GRU~\cite{xu2021limu} & 0.921 & 0.927 & 0.965 & 0.969 \\
GRU~\cite{chung2014empirical} & 0.921 & 0.930 & 0.965 & 0.971 \\
DCNN~\cite{yang2015deep} & 0.845 & 0.844 & 0.915 & 0.911 \\
Deepsense~\cite{yao2017deepsense} & 0.667 & 0.660 & 0.614 & 0.588 \\
LSTM~\cite{shi2015convolutional} & 0.877 & 0.899 & 0.939 & 0.948 \\
\bottomrule 
\end{tabular}
}
\end{table}

\begin{figure}
    \centering
    \includegraphics[width=0.9\linewidth]{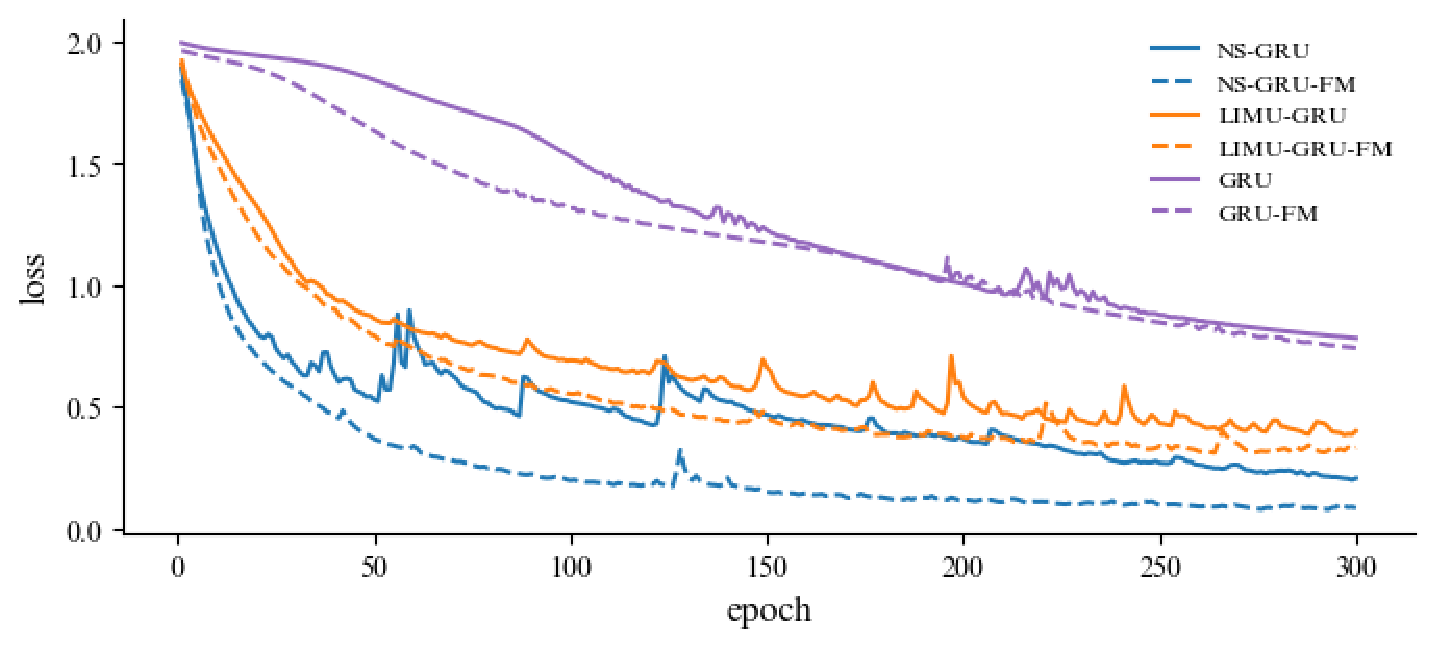}
    \caption{The training loss with epochs.}
    \label{fig:loss-epoch}
\end{figure}

\subsubsection{Loss Comparison}
For training the classifier on mobile devices, faster convergence will reduce the time and power consumption. Figure~\ref{fig:loss-epoch} shows the training loss of different models on common activities of OPPOHAR dataset with epochs. Results show our NS-GRU-FM model converges fastest, indicating the effectiveness of learning better representation of our network architecture and data augmentation method. This ensures the possibility of finetuning the classifier on user devices.



\section{Conclusion}
In this paper, we present a smartphone-IMU-based human activity recognition dataset named OPPOHAR. 
To enable the user-specific activity detection and to promote the implementation in mobile devices, we propose Non-stationary BERT network and a new data augmentation method for IMU data. 
Experiment results show the effectiveness of NS-BERT model and the universality of the data augmentation method. 

\bibliographystyle{unsrt}  
\bibliography{references}  

\footnote{This paper is submitted to Interspeech2025.}

\end{document}